\pdfoutput=1

\documentclass[11pt]{article}

\usepackage{emnlp2021}

\usepackage{times}
\usepackage{latexsym}
\usepackage{amsmath}
\usepackage{mathtools}
\usepackage{amssymb}
\usepackage{array}
\usepackage{amsthm}
\usepackage{multirow}
\usepackage{bbold}
\usepackage{cleveref}
\usepackage{subcaption}
\usepackage{booktabs}
\usepackage{enumitem}
\usepackage{graphicx}

\usepackage[T1]{fontenc}

\usepackage[utf8]{inputenc}

\usepackage{microtype}

\title{Non-Parametric Unsupervised Domain Adaptation for Neural \\ Machine Translation}

\author{
  Xin Zheng$^1$, Zhirui Zhang$^2$,  Shujian Huang$^{13}$\thanks{~ ~Corresponding author.}, Boxing Chen$^2$, Jun Xie$^2$, \\ 
  {\bf Weihua Luo$^2$ and Jiajun Chen$^1$} \\
  $^1$National Key Laboratory for Novel Software Technology, Nanjing University, China \\
  $^2$Language Technology Lab, Alibaba DAMO Academy \;
  $^3$Peng Cheng Laboratory, China \\
  $^1$\texttt{zhengxin@smail.nju.edu.cn},\; \texttt{\{huangsj,chenjj\}@nju.edu.cn}\\ 
  $^2$\texttt{\{boxing.cbx, qingxing.xj, weihua.luowh\}@alibaba-inc.com} \\
  $^2$\texttt{zrustc11@gmail.com}
}

\begin{document}
\maketitle
\begin{abstract}

Recently, $k$NN-MT \citep{DBLP:journals/corr/abs-2010-00710} has shown the promising capability of directly incorporating the pre-trained neural machine translation (NMT) model with domain-specific token-level $k$-nearest-neighbor ($k$NN) retrieval to achieve domain adaptation without retraining.
Despite being conceptually attractive, it heavily relies on high-quality in-domain parallel corpora, limiting its capability on unsupervised domain adaptation, where in-domain parallel corpora are scarce or nonexistent.
In this paper, we propose a novel framework that directly uses in-domain monolingual sentences in the target language to construct an effective datastore for $k$-nearest-neighbor retrieval.
To this end, we first introduce an autoencoder task based on the target language, and then insert lightweight adapters into the original NMT model to map the token-level representation of this task to the ideal representation of translation task.
Experiments on multi-domain datasets demonstrate that our proposed approach significantly improves the translation accuracy with target-side monolingual data, while achieving comparable performance with back-translation.
Our implementation is open-sourced at \url{https://github.com/zhengxxn/UDA-KNN}.

\end{abstract}

\section{Introduction}

Non-parametric methods \cite{DBLP:conf/aaai/GuWCL18, zhang-etal-2018-guiding, bapna-firat-2019-non, DBLP:journals/corr/abs-2010-00710, zheng-etal-2021-adaptive} have recently been successfully applied to neural machine translation (NMT).
These approaches complement advanced NMT models \cite{DBLP:conf/nips/SutskeverVL14, DBLP:journals/corr/BahdanauCB14, DBLP:conf/nips/VaswaniSPUJGKP17, DBLP:journals/corr/abs-1803-05567} with external memory to alleviate the performance degradation when translating out-of-domain sentences, rare words~\cite{DBLP:conf/aclnmt/KoehnK17}, etc.
Among them, $k$NN-MT~\cite{DBLP:journals/corr/abs-2010-00710} is a simple yet effective non-parametric method using nearest neighbor retrieval.
More specifically, $k$NN-MT equips a pre-trained NMT model with a $k$NN classifier over a provided datastore of cached context representations and corresponding target tokens to improve translation accuracy without retraining.
This promising ability to access any provided datastore or external knowledge during inference makes it expressive, adaptable, and interpretable.

Despite the potential benefits, $k$NN-MT requires large-scale in-domain parallel corpora to achieve domain adaptation. 
However, in practice, it is not realistic to collect large amounts of high-quality parallel data in every domain we are interested in.
Since monolingual in-domain data is usually abundant and easy to obtain, it is essential to explore the capability of $k$NN-MT on unsupervised domain adaptation scenario that utilizes large amounts of monolingual in-domain data.
One straightforward and effective solution for unsupervised domain adaptation is to build in-domain synthetic parallel data via back-translation of monolingual target sentences \cite{sennrich-etal-2016-improving, DBLP:journals/corr/abs-1803-00353, dou-etal-2019-unsupervised,wei-etal-2020-iterative}.
Although this approach has proven superior effectiveness in exploiting monolingual data, applying it in $k$NN-MT requires an additional reverse model and brings the extra cost of generating back-translation, making the adaptation of $k$NN-MT more complicated and time-consuming in practice.

In this paper, we propose a novel \textbf{U}nsupervised \textbf{D}omain \textbf{A}daptation framework based on $k$NN-MT (UDA-$k$NN).
The UDA-$k$NN aims at directly leveraging the monolingual target-side data to generate the corresponding datastore, and encouraging it to play a similar role with the real bilingual in-domain data, through the carefully designed architecture and loss function. 
Specifically, we introduce an autoencoder task based on target language to enable datastore construction with monolingual data.
Then we incorporate lightweight adapters into the encoder part of pre-trained NMT model to make the decoder's representation in autoencoder task close to the corresponding representation in translation task. 
In this way, the adapter module implicitly learns the semantic mapping from the target language to source language in feature space to construct an effective in-domain datastore, while saving the extra cost of generating synthetic data via back-translation.

We evaluate the proposed approach on multi-domain datasets, including IT, Medical, Koran and Law domains.
Experimental results show that when using target-side monolingual data, our proposed approach obtains $4.9$ BLEU improvements on average and even achieves similar performance compared with back-translation.

\section{Background}
In this section, we give a brief introduction to the domain adaptation of $k$NN-MT. In general, the process includes two steps: creating an in-domain datastore and decoding with retrieval on it.

\paragraph{In-domain Datastore Creation.} 
Given a pre-trained general domain NMT model and an in-domain parallel corpus, $k$NN-MT utilizes the model to forward pass the corpus to create a datastore.
Formally, for each bilingual sentence pair in the corpus $(x, y) \in (\mathcal{X}, \mathcal{Y})$, the NMT model will generate a context representation $h(x, y_{<t})$ for each target-side token $y_t$. 
Then, the datastore is constructed by collecting the representations and corresponding tokens as keys and values respectively:
\begin{equation}
(\mathcal{K}, \mathcal{V}) = \bigcup_{(x, y) \in (\mathcal{X}, \mathcal{Y})} \{(h(x, y_{<t}), y_t), \forall y_t \in y \}. 
\label{equ:datastore}
\end{equation}
\paragraph{Decoding with Retrieval.} 
On each decoding step $t$, the NMT model first generates a representation $h(x, \hat{y}_{<t})$ for current translation context, which consists of source-side $x$ and generated target-side tokens $\hat{y}_{<t}$.
Then, the representation is used to query the in-domain datastore for $k$ nearest neighbors, which can be denoted as $N = \{(h_i, v_i), i \in \{1, 2, ..., k\} \}$.
These neighbors are utilized to form a distribution over the vocab:
\begin{align}
p_{\textrm{kNN}}(y_t & |x, \hat{y}_{<t}) \propto \label{equ:knn_prob} \\
&\sum_{(h_i, v_i)} \mathbb{1}_{y_t = v_i} \exp (\frac{-d(h_i, h(x, \hat{y}_{<t}))}{T}), \nonumber 
\end{align}
where $T$ is the temperature and $d(\cdot, \cdot)$ indicates the squared euclidean distance.
The final probability to predict next token $y_{t}$ is an interpolation of two distributions with a hyper-parameter $\lambda$:
\begin{equation}
\label{equ:prob_ip}
\begin{split}
p(y_t|x, \hat{y}_{<t}) &= \lambda \ p_{\textrm{kNN}}(y_t|x, \hat{y}_{<t}) \\
& + (1-\lambda) \ p_{\textrm{NMT}} (y_t|x, \hat{y}_{<t}), 
\end{split}
\end{equation}
where $p_{\textrm{NMT}}$ indicates the general domain NMT prediction and $p_{\textrm{kNN}}$ represents the in-domain retrieval based prediction. 
 

 
    
\begin{figure}[t]
    \centering
    \includegraphics[width=0.49\textwidth]{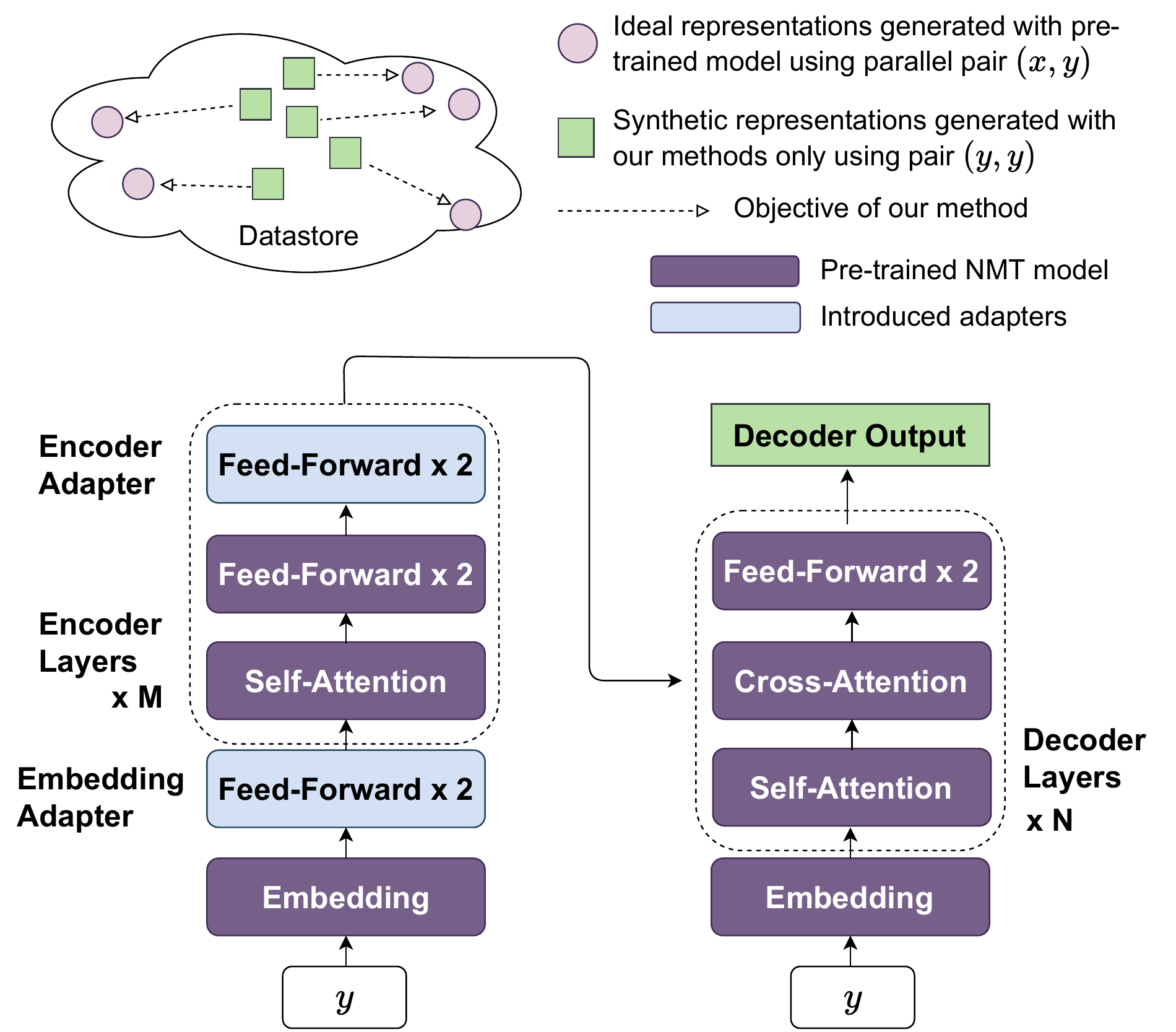}
    \vspace{-10pt}
    \caption{An overview of the proposed method.}
    \label{fig:model}
\end{figure}

\section{Unsupervised Domain Adaptation with $k$NN-MT}

Although \citet{DBLP:journals/corr/abs-2010-00710} has shown the capability of $k$NN-MT on domain adaptation, the datastore creation heavily relies on high-quality in-domain parallel data, which cannot be always satisfied in practice. 
As in-domain monolingual data is usually abundant and easy to obtain, it is essential to extend the capability of $k$NN-MT on unsupervised domain adaptation that merely uses large amounts of in-domain target sentences.
In this paper, we design a novel non-parametric \textbf{U}nsupervised \textbf{D}omain \textbf{A}daptation framework based on $k$NN-MT (UDA-$k$NN) to fully leverage in-domain target-side monolingual data.

The overview framework of UDA-$k$NN is illustrated in Figure \ref{fig:model}. 
The UDA-$k$NN starts with the autoencoder task based on target language $y$, where the target-side is simply copied to the source-side to generate pair $(y, y)$.
Based on that, the UDA-$k$NN aims to make the decoder's representation in autoencoder task close to the ideal representation in translation task.
In this way, we can directly leverage autoencoder structure and in-domain target sentences to construct the corresponding datastore for $k$-nearest-neighbor retrieval, which is similar to that from real in-domain bilingual data.
Next, we will introduce the architecture and training objective of our proposed method in detail.




\paragraph{Architecture.} We insert lightweight adapter layers~\cite{DBLP:conf/icml/HoulsbyGJMLGAG19,DBLP:conf/nips/GuoZXWCC20,DBLP:journals/taslp/GuoZXCC21} into the source embedding layer and each encoder layer of the pre-trained NMT model to perform the autoencoder task, by which we only increase a few parameters for our method.
Specifically, we simply construct the adapter layer with layer normalization as well as two feed-forward networks with non-linearity between them:
\begin{equation}
\label{equ:enc_adapter_forward}
Z = W_1 \cdot (\textrm{LN}(H)), \\
H_{\textsc{o}} = H + W_2 \cdot (\textrm{ReLU}(Z)),
\end{equation}
where $H$ and $H_{\textsc{o}}$ are the input and output hidden states of the adapter layer respectively, $\textrm{LN}$ indicates layer normalization, $W_1$ and $W_2$ are the parameters of the feed-forward networks. 

\paragraph{Training.}
The UDA-$k$NN is designed to leverage monolingual target-side data to generate the corresponding datastore, which plays a similar role with real in-domain bilingual data.
We achieve this by leveraging out-of-domain bilingual data $(\mathcal{X}, \mathcal{Y})$.
More specifically, given a bilingual sentence pair in the corpus $(x, y) \in (\mathcal{X}, \mathcal{Y})$, the original NMT model generates decoder representation $h_{(x;\ y_{<t})}$ for each target token $y_t$.
Meanwhile, with the target-copied pair $(y, y)$, the NMT model with adapters generates another representation for each $y_t$, which can be denoted as $h'_{(y;\ y_{<t})}$.
We take the end-to-end paradigm to directly optimize the adapter layers by minimizing the squared euclidean distance of the two sets of decoder representations:
\begin{equation}
\theta^* = \min_{\theta} \sum_{(x,\ y) \in (\mathcal{X}, \mathcal{Y})} \sum_{t}||h'_{(y;\ y_{<t})} - h_{(x;\ y_{<t})}||^2,
\label{equ:objective}
\end{equation}
where $\theta$ is the parameters of all adapter layers.  
Note that we keep original parameters in the pre-trained NMT model fixed during training to avoid the performance degradation of the NMT model in the inference stage.

\paragraph{Prediction.}
For unsupervised domain adaptation, given the domain-specific target-side monolingual data, we first copy the target sentences to the source side to generate synthetic bilingual pairs.
Then the pre-trained NMT model with adapter layers forward passes these pairs to create an in-domain datastore.
When translating in-domain sentences, we utilize the original NMT model and $k$NN retrieval on the in-domain datastore to perform online domain adaptation as \Cref{equ:prob_ip}.

\section{Experiments}

\subsection{Setup}
\paragraph{Datasets and Evaluation Metric.} 
We use the same multi-domain dataset as ~\citet{aharoni-goldberg-2020-unsupervised} to evaluate the effectiveness of our proposed model and consider domains including \textbf{IT}, \textbf{Medical}, \textbf{Koran}, and \textbf{Law} in our experiments.
We extract target-side data in the training sets to perform unsupervised domain adaptation while keeping the dev and test sets unchanged.
Besides, WMT'19 News data\footnote{\url{http://www.statmt.org/wmt19/translation-task.html}} is used for training the adapters in our method as well as the reverse translation model for back-translation.
The sentence statistics of all datasets are illustrated in \cref{table:dataset}.
The Moses toolkit\footnote{\url{https://github.com/moses-smt/mosesdecoder}} is used to tokenize the sentences and we split the words into subword units \citep{sennrich-etal-2016-neural} with the codes provided by the pre-trained model \citep{ng-etal-2019-facebook}.
We use SacreBLEU\footnote{\url{https://github.com/mjpost/sacrebleu}} to measure all results with case-sensitive detokenized BLEU~\citep{papineni-etal-2002-bleu}.

\begin{table}[htb] \small
\centering
\resizebox{0.48\textwidth}{!}{%
\begin{tabular}{l|ccccc}
\toprule
\multicolumn{1}{l|}{Dataset} & WMT19'News & IT    & Medical & Koran & Laws    \\ \midrule
Train                     &$37,079,168$ & $222,927$ & $248,009$   & $17,982$ & $467,309$ \\ 
Dev            &$10,000$ & $2000$& $2000$   & $2000$ & $2000$ \\ 
Test            &  -       & $2000$ & $2000$   & $2000$ & $2000$ \\ \bottomrule
\end{tabular}%
}
\vspace{-5pt}
\caption{Statistics of dataset in different domains.}
\label{table:dataset}
\vspace{-5pt}
\end{table}


\paragraph{Methods.}
We compare our proposed approach with several baselines:
\begin{itemize}[leftmargin=*]
    \setlength{\itemsep}{0pt}
    \setlength{\parskip}{0pt}
    \item \textbf{Basic NMT}: A general domain model is directly used to evaluate on in-domain test sets.
    \item \textbf{Empty-$k$NN}: The source-side of synthetic bilingual data is always set to \textit{<EOS>} token. 
    \item \textbf{Copy-$k$NN}: Each target sentence is copied to source-side to produce synthetic bilingual data. This is a special case of our method without model training.
    \item \textbf{BT-$k$NN}: A reverse translation model is applied to produce synthetic bilingual data, which are used to generate in-domain datastore.
    \item \textbf{Parallel-$k$NN}: Ground-truth parallel data is used to generate the in-domain datastore, which can be regarded as the upper bound of the $k$NN retrieval based methods.
\end{itemize}

\paragraph{Implementation Details.}

We use the WMT'19 German-English News translation task winner model \citep{ng-etal-2019-facebook} as our general domain model.
For introduced adapters, the hidden size is set to $1024$, with only about $6\%$ parameters of the original model.
Adam \citep{DBLP:journals/corr/KingmaB14} is used to update the parameters in adapters. 
During training, we collect about $40000$ tokens for each batch and schedule the learning rate with the inverse square root decay scheme, in which the warm-up step is set as $4000$, and the maximum learning rate is set as 7e-4.
Faiss\footnote{\url{https://github.com/facebookresearch/faiss}} is used to build the in-domain datastore to carry out fast nearest neighbor search. We utilize faiss to learn $4096$ cluster centroids for each domain, and search $32$ clusters for each target token in decoding.
When inference, we retrieve $16$ nearest neighbors in the datastore. 
We set the hyper-parameter $T$ as $4$ for IT, Medical, Law, and $40$ for Koran.
The $\lambda$ is tuned on the in-domain dev sets for different methods.

\subsection{Main Results}

The adaptation performance of different methods are listed in \Cref{table:main}.
Obviously, our method can significantly improve the translation accuracy on in-domain test sets compared to basic NMT, while Empty-$k$NN and Copy-$k$NN can't.
It demonstrates the efficiency of our proposed method to create an in-domain datastore by leveraging only monolingual data.
Besides, we can observe that our method achieves comparable performance over BT-$k$NN,
but completely avoids the reverse model and extra time cost to generate synthetic data, making the adaptation much faster and simpler.

\begin{table}[tb]
\centering \small
\setlength\tabcolsep{3pt}
\begin{tabular}{@{}l|ccccc}
\toprule
Model     & IT    & Medical & Law   & Koran & Avg.   \\ \midrule
Basic NMT  & $38.35$ & $39.99$   & $45.48$ & $16.26$ & $35.02$                        \\ \midrule
Empty-$k$NN & $38.06$   & $40.01$     & $45.62$   & $16.44$   & $35.03$                          \\
Copy-$k$NN  & $38.96$ & $40.86$   & $46.00$ & $17.06$ & $35.72$                  \\
BT-$k$NN    & $41.35$ & $47.02$   & $52.91$ & $19.58$ & $40.23$                          \\
UDA-$k$NN      & $41.57$ & $46.64$   & $52.02$ & $19.42$ & $39.91$                        \\ \midrule
Parallel-$k$NN &
  \multicolumn{1}{c}{$45.96$} &
  \multicolumn{1}{c}{$54.16$} &
  \multicolumn{1}{c}{$61.31$} &
  \multicolumn{1}{c}{$20.30$} &
  \multicolumn{1}{c}{$45.43$} \\ \bottomrule
\end{tabular}%
\vspace{-5pt}
\caption{The BLEU scores [\%] of different methods evaluated on four domains.}
\label{table:main}
\vspace{-5pt}
\end{table}

\subsection{Analysis}

In this section, we would like to further explore the reasons behind the success of our approach.

\paragraph{Similarity Measurement.} 
We measure the cosine similarity and squared euclidean distances between the synthetic representations generated by our method and ideals generated using ground-truth parallel data. 
As shown in \Cref{table:similarity}, we also list the results of BT-$k$NN and Copy-$k$NN.
We can observe that even without the source language information, our UDA-$k$NN can generate the representations that are close enough to the ideals as BT-$k$NN, leading to the efficient in-domain retrieval for $k$NN-MT.
It also verifies the effectiveness of the adapter layers on directly learning the semantic mapping from target language to source language in feature space.

\begin{table}[htb]
\centering \small
\setlength\tabcolsep{3pt}
\begin{tabular}{@{}lcccc@{}}
\toprule
\multicolumn{5}{c}{Cosine Similarity ($\uparrow$)}                               \\ \midrule
\multicolumn{1}{l|}{Method}   & IT    & Medical & Law  & Koran   \\ \midrule
\multicolumn{1}{l|}{Copy-$k$NN} & $0.74$  & $0.77$    & $0.77$ & $0.65$  \\
\multicolumn{1}{l|}{BT-$k$NN}   & $0.85$  & $0.86$   & $0.92$  & $0.81$  \\
\multicolumn{1}{l|}{UDA-$k$NN}     & $0.87$  & $0.87$    & $0.91$ & $0.84$ \\ \midrule
\multicolumn{5}{c}{Squared Euclidean Distance ($\downarrow$)}                   \\ \midrule
\multicolumn{1}{l|}{Method}   & IT    & Medical & Law  & Koran   \\ \midrule
\multicolumn{1}{l|}{Copy-$k$NN} & $85.93$ & $79.03$   & $77.97$ & $145.93$ \\
\multicolumn{1}{l|}{BT-$k$NN}   & $47.00$ & $42.30$   & $25.47$ & $78.77$  \\
\multicolumn{1}{l|}{UDA-$k$NN}     & $46.10$ & $45.83$   & $31.36$ & $68.56$ \\ \bottomrule
\end{tabular}%
\vspace{-5pt}
\caption{ Cosine similarity / squared euclidean distance between the ground-truth representations and that generated by different methods.}
\label{table:similarity}
\vspace{-5pt}
\end{table}

\paragraph{Visualization.}

\begin{figure*}[t]
    \centering
    \includegraphics[width=0.95\textwidth]{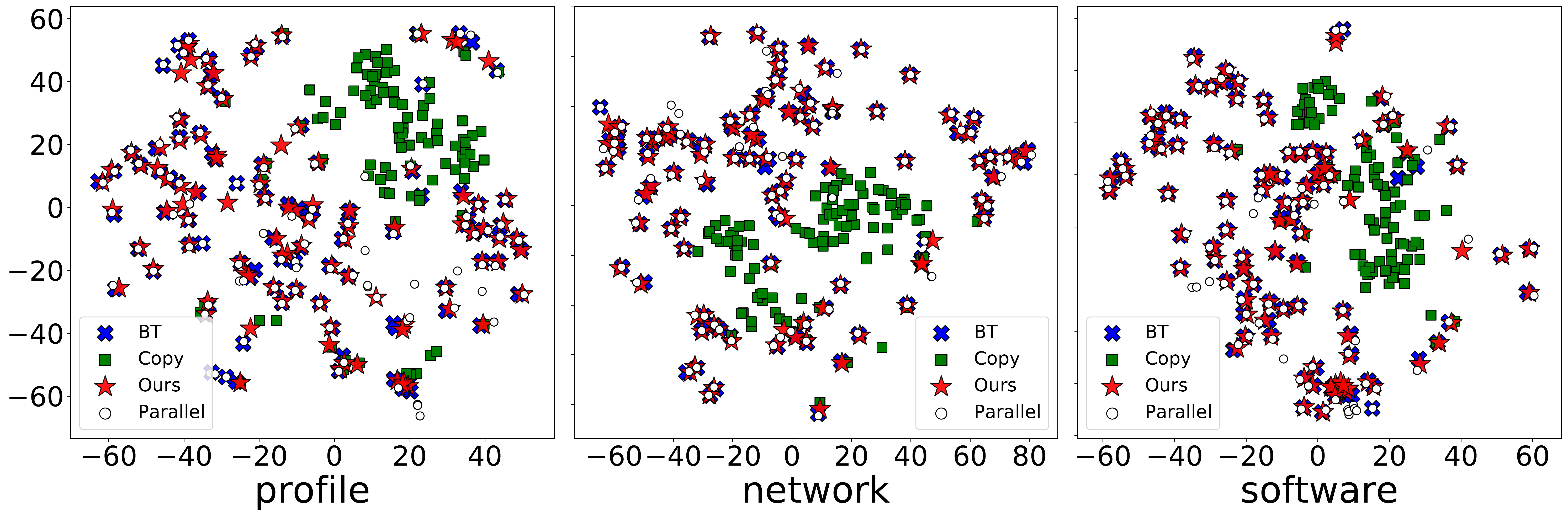}
    \vspace{-5pt}
    \caption{The t-SNE visualization \citep{vanDerMaaten2008} of the representation distributions for label \textit{profile}, \textit{network}, \textit{software} in the datastores, which are created by different methods.} 
    \label{fig:visualization}
    \vspace{-8pt}
\end{figure*}

We also collect and visualize the representations with the same target tokens in different datastores to give intuitive insights of the impact of adapters.
Specifically, we select three common words in IT domain and show the results in \Cref{fig:visualization}.
We can see that the representations generated with Copy-$k$NN tend to gather in small areas, which results in retrieval collapse when meeting diverse translation contexts.
While with the adapters, the distribution of the same label in the datastore can be closer to that generated with bilingual pairs, improving the retrieval efficiency. 

\paragraph{Effect of Adapter Position.}

In our proposed method, we only insert adapters into the encoder side as we would like to modify the encoding function of $y$. 
It aims to encode the $y$ into the same feature space as the semantically identical $x$.
We also compare our choice to the common practice \citep{ bapna-firat-2019-simple, DBLP:conf/nips/GuoZXWCC20}, where the adapters are inserted into both encoder and decoder sides.
The results are shown in \Cref{table:effect}.
We can observe that the adapters in the decoder side can only play a very limited role, which also demonstrates the motivation of our approach.

\begin{table}[htb]
\centering \small
\setlength\tabcolsep{3pt}
\begin{tabular}{@{}l|ccccc@{}}
\toprule
Adapter             & IT    & Medical & Law & Koran   & Avg   \\ \midrule
Encoder          & $41.57$ & $46.64$   & $52.02$ & $19.42$ & $39.91$ \\
Encoder + Decoder & $41.73$ & $46.75$   & $52.15$ & $19.31$ & $39.99$ \\ \bottomrule
\end{tabular}%
\vspace{-5pt}
\caption{The BLEU scores [\%] of inserting adapters into encoder / encoder and decoder sides of our method.}
\label{table:effect}
\end{table}

\paragraph{Comparison with Fine-tuning Strategy.}
We compare our method with BT-FT, where the back-translation data is used for fine-tuning the full NMT model. 
The fine-tuning method easily causes catastrophic forgetting problem \citep{thompson-etal-2019-overcoming} and results in performance degradation, especially when the data contains noise, as the results shown in \Cref{table:FT}.

\begin{table}[htb]
\centering \small
\setlength\tabcolsep{3pt}
\begin{tabular}{@{}l|ccccc@{}}
\toprule
Method             & IT    & Medical & Law & Koran   & Avg   \\ \midrule
Basic NMT  & $38.35$ & $39.99$   & $45.48$ & $16.26$ & $35.02$                        \\ \midrule
UDA-$k$NN          & $41.57$ & $46.64$   & $52.02$ & $19.42$ & $39.91$ \\
BT-$k$NN    & $41.35$ & $47.02$   & $52.91$ & $19.58$ & $40.23$    \\
BT-FT              & $39.72$ & $46.44$   & $51.06$ & $17.45$ & $38.67$ \\ \bottomrule
\end{tabular}%
\vspace{-5pt}
\caption{The BLEU scores [\%] of the non-parametric methods and fine-tuning method.}
\label{table:FT}
\end{table}

\section{Conclusion}

In this paper, we present UDA-$k$NN, a simple yet effective framework that directly utilizes monolingual data to construct in-domain datastore for unsupervised domain adaptation of $k$NN-MT.
Experimental results verify that our method obtains significant improvement with target-side monolingual data.
Our approach also achieves comparable performance with the BT-based method, while saving the extra cost of generating back-translation.

\section{Acknowledgments}
We would like to thank the anonymous reviewers for their insightful comments.
This work is supported by National Science Foundation of China (No. U1836221, 61772261), National Key R\&D Program of China (No. 2019QY1806) and Alibaba Group through Alibaba Innovative Research Program.
We appreciate Weizhi Wang, Hao-Ran Wei, Yichao Du for the fruitful discussions. 
The work was done when the first author was an intern at Alibaba Group.

\bibliography{anthology,custom}
\bibliographystyle{acl_natbib}

\appendix

\section{Appendix}




\subsection{Translation Examples}

\Cref{table:case_study} shows a translation example selected from the Medical dataset.
We can observe that our proposed UDA-$k$NN can make more proper word selections compared with basic NMT as well as Copy-$k$NN, thanks to the effective in-domain datastore construction.
In addition, the overall translation accuracy of our method is close to BT-$k$NN and parallel-$k$NN, which utilize bilingual pairs to create datatsore while we only use monolingual pairs.

\begin{table*}[t] \small
\centering
\begin{tabular}{@{}r|l@{}}
\toprule
Source           & Insbesondere bei großen chirurgischen Eingriffen ist eine genaue Überwachung der Substitutionstherapie \\ & mit Hilfe einer Koagulationsanalyse (Faktor VIII-Aktivität im Plasma) unbedingt erforderlich. \\ \midrule
Reference      & In the case of major surgical interventions in particular, precise monitoring of the substitution therapy \\ & by means of coagulation analysis (plasma factor VIII activity) is indispensable. \\ \midrule
Basic NMT       &  Particularly in the case of major surgical procedures, precise monitoring of the substitution therapy \\ & with the help of a coagulation analysis (factor VIII activity in the plasma) is absolutely necessary.  \\ \midrule
Copy-$k$NN      &  Particularly in case of major surgical \textbf{\textit{interventions}} a precise monitoring of the substitution therapy \\ & with the help of a coagulation analysis (factor VIII activity in the plasma) is necessary.  \\ \midrule
BT-$k$NN         & In the case of major surgical \textbf{\textit{interventions}} in particular, precise monitoring of the substitution therapy \\ & \textbf{\textit{by means of}} a coagulation analysis (factor VIII activity in plasma) is \textbf{\textit{indispensable}}.  \\ \midrule
UDA-$k$NN        & In the case of major surgical \textbf{\textit{interventions}} in particular, precise monitoring of the substitution therapy \\ & \textbf{\textit{by means of}} coagulation analysis (factor VIII activity in the plasma) is \textbf{\textit{indispensable}}.  \\ \midrule
Parallel-$k$NN & In the case of major surgical \textbf{\textit{interventions}} in particular, precise monitoring of the substitution therapy \\ & \textbf{\textit{by means of}} a coagulation analysis (plasma factor VIII activity) is \textbf{\textit{indispensable}}. \\ \bottomrule
\end{tabular}%
\vspace{-5pt}
\caption{Translation examples of different systems in Medical domain.}
\label{table:case_study}
\vspace{-5pt}
\end{table*}

\end{document}